\documentclass{article}
\usepackage{iclr2027_conference,times}

\usepackage[T1]{fontenc}
\usepackage[utf8]{inputenc}
\usepackage{microtype}
\usepackage{booktabs}
\usepackage{tabularx}
\usepackage{array}
\usepackage{enumitem}
\usepackage{graphicx}
\usepackage{xcolor}
\usepackage{caption}
\usepackage{todonotes}  
\usepackage{xurl}
\definecolor{hzBuilder}{HTML}{2A78D6}
\definecolor{hzReviewer}{HTML}{EB6834}
\definecolor{hzHuman}{HTML}{1BAF7A}
\definecolor{hzCommunity}{HTML}{4A3AA7}
\definecolor{hzGood}{HTML}{0CA30C}
\definecolor{hzWarning}{HTML}{FAB219}
\definecolor{hzSerious}{HTML}{EC835A}
\definecolor{hzCritical}{HTML}{D03B3B}
\definecolor{hzInk}{HTML}{0B0B0B}
\definecolor{hzInkSecondary}{HTML}{52514E}
\definecolor{hzMuted}{HTML}{898781}
\definecolor{hzAxis}{HTML}{C3C2B7}
\definecolor{hzGrid}{HTML}{E1E0D9}
\definecolor{hzWash}{HTML}{F0EFEC}
\definecolor{hzBlue}{HTML}{2A78D6}
\definecolor{hzOrange}{HTML}{EB6834}
\definecolor{hzAqua}{HTML}{1BAF7A}
\definecolor{hzYellow}{HTML}{EDA100}
\definecolor{hzMagenta}{HTML}{E87BA4}
\definecolor{hzGreen}{HTML}{008300}
\definecolor{hzViolet}{HTML}{4A3AA7}
\definecolor{hzRed}{HTML}{E34948}
\definecolor{hzSeq0}{HTML}{CDE2FB}
\definecolor{hzSeq1}{HTML}{B7D3F6}
\definecolor{hzSeq2}{HTML}{9EC5F4}
\definecolor{hzSeq3}{HTML}{86B6EF}
\definecolor{hzSeq4}{HTML}{6DA7EC}
\definecolor{hzSeq5}{HTML}{5598E7}
\definecolor{hzSeq6}{HTML}{3987E5}
\definecolor{hzSeq7}{HTML}{2A78D6}
\definecolor{hzSeq8}{HTML}{256ABF}
\definecolor{hzSeq9}{HTML}{1C5CAB}
\definecolor{hzSeq10}{HTML}{184F95}
\definecolor{hzSeq11}{HTML}{104281}
\definecolor{hzSeq12}{HTML}{0D366B}

\usepackage[colorlinks=true,citecolor=hzBuilder,linkcolor=hzInk,urlcolor=hzViolet]{hyperref}

\setlist{leftmargin=*,nosep}
\newcommand{\system}{Humanize}

\title{Humanize: Judgement Engineering for\\Agentic Coding}
\author{Sihao Liu\textsuperscript{1,2}\thanks{Work done while at UCLA.},
 Ligeng Zhu\textsuperscript{1}, Zijian Zhang\textsuperscript{1,4},
Dongyun Zou\textsuperscript{3,5}, Zhengyang Zhang\textsuperscript{4} \\
\textbf{
Changye Li\textsuperscript{4},
Song Bian\textsuperscript{1},
Song Han\textsuperscript{1,3},
Tony Nowatzki\textsuperscript{2}
}
\\[0.5ex]
\normalfont\small
\textsuperscript{1}NVIDIA \quad \textsuperscript{2}UCLA \quad
\textsuperscript{3}MIT \quad \textsuperscript{4}Tsinghua University \quad \textsuperscript{5}Nunchux AI}
\date{}
\iclrfinalcopy

\begin{document}
\maketitle
\lhead{Preprint}
\begin{abstract}
Agentic coding makes code generation cheap, but reliable completion remains difficult: the agent that writes
the code is a weak judge of whether it is done. We present \system{}, a multi-agent orchestration workflow
for agentic coding built around \emph{judgement engineering}: explicit, mechanically enforced decisions at the
boundaries between planning, implementation, review, and learning. A human approves a plan contract, a builder
agent implements it in rounds, and a reviewer agent from another vendor decides completion; deterministic
hooks, not a model, route work between these roles and enforce 72 mechanical gates. Viewed as a Markov chain
over repository states, alternating builder and reviewer samples jointly from two models, so a defect survives
only if both miss it. We study \system{}
through its deployment, 118 public postmortems of real loops, and its applications. Over 68 versions in 108
days, it gathered 1,468 GitHub stars. Applications include a 567-file gem5 build-system migration under
upstream review; Kernel Design Agents, which extend the loop with a kernel knowledge base and profiling feedback
and placed in the top three of all three Full-Agent tracks of the MLSys 2026 FlashInfer contest; and, through
Humanize Olympiad Agents (HOA), full scores in IOI 2026, IMO 2026, IPhO 2026, and IBO 2024, 418.5/437 in IChO 2026 (gold-medal).
\system{} also achieves 672/672 on PutnamBench and ranks first (251/303) on Lean-Eval's
leaderboard even competiting with professional mathematicians. The postmortems show that independent review catches unsupported builder claims, but
stopping remains a key weakness. In reports that separate rounds by phase, two thirds of rounds occurred
after implementation was accepted. This evidence is observational, not a controlled comparison of workflows. \\\\
Code:\footnote{Meanwhile, Humanize2 is in active development at \url{https://github.com/humanfia/humanize}.} \url{https://github.com/PolyArch/humanize}
\end{abstract}

\section{Introduction}
\label{sec:intro}

Agentic coding makes writing code cheap. Instead of stopping after a single prompt, agents such as Claude Code
and Codex CLI pursue a target through cycles of planning, implementation, review, and
revision~\citep{anthropic2025claudecode,openai2025codexcli}. Errors accumulate as the implementation grows, and
the agent that wrote the code is a weak judge of it: language models struggle to correct their own reasoning
without external feedback~\citep{huang2024cannotselfcorrect}, favor their own
outputs~\citep{panickssery2024selfpreference}, and can reward-hack by changing
tests~\citep{vonarx2025rewardhacking}. The challenge shifts from producing code to verifying correctness, goal
alignment, and completion.

Existing designs leave the completion decision with the agents that do the work. Self-refinement loops let a
model revise its own output~\citep{shinn2023reflexion,madaan2023selfrefine}, task verification is a major
failure category of multi-agent systems~\citep{cemri2025mast}, and the Ralph loop re-feeds a coding agent the
same prompt until the agent itself declares the task done~\citep{huntley2025ralph,anthropic2025ralphwiggum}.
Reliable completion instead requires explicit decisions about the plan, the builder's claims, the review, and
when to stop.

\begin{figure}[t]
  \centering
  \includegraphics[width=\linewidth]{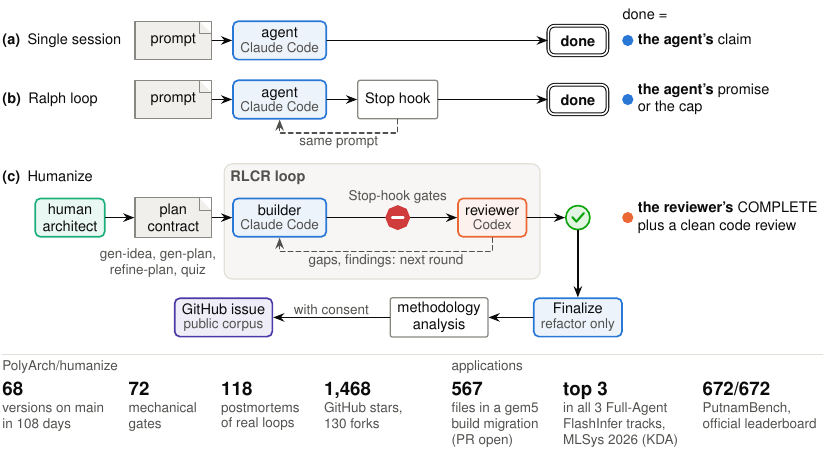}
  \caption{\textbf{\system{} separates building from acceptance through independent review and mechanical gates.}
    (a)--(c)~The right column compares who declares completion in a single session, a Ralph
    loop~\citep{huntley2025ralph}, and \system{}. In (c), Stop-hook gates and an independent reviewer check each
    builder round; dashed arrows return work. The bottom strip summarizes the project's development record
    and applications.}
  \label{fig:overview}
\end{figure}

We present \system{}, a multi-agent orchestration workflow built around
\emph{judgement engineering}: explicit, mechanically enforced decisions at the boundaries between planning,
implementation, review, and learning (Figure~\ref{fig:overview}). The human approves a plan contract that fixes
the goal and its acceptance criteria. A builder agent implements the plan in rounds and summarizes its claims. A
reviewer agent from another vendor evaluates alignment and correctness and alone can declare the work complete.
Deterministic hooks, not a model, route work between these roles and enforce mechanical gates, so the roles
hold when an agent errs or over-claims. Viewed as a Markov chain over repository states, the alternation of
builder and reviewer samples jointly from two models, and a defect survives only if both miss it.

We study \system{}, a Claude Code plugin whose core loop is RLCR (Ralph-Loop with Codex Review). Our
contributions are:
\begin{itemize}
\item \textbf{Design.} Judgement engineering for orchestrating coding agents, built on alternating two models
  as a joint sampler and realized in a plan contract, a round protocol between builder and reviewer, and 72
  mechanical gates (\S\ref{sec:judgement} and~\S\ref{sec:system}).
\item \textbf{Field evidence.} A study of the deployment of \system{} and of 118 public postmortems of real
  loops, most written by its own methodology analysis (\S\ref{sec:deployment}).
\item \textbf{Applications and lessons.} Applications to a mapper optimization, a 567-file build migration,
  contest GPU kernels from a loop extended with a kernel knowledge base and profiling feedback, PutnamBench, and
  olympiads, which show that independent judgement can come from a second model or from an oracle
  (\S\ref{sec:applications}), and five design lessons (\S\ref{sec:findings}).
\end{itemize}
The evidence is observational: it shows what the design enables and where it fails, not how it compares with
other workflows under controlled conditions.

\section{Judgement Engineering}
\label{sec:judgement}

\system{} starts from one premise: \textbf{do not automate a plan the human cannot judge.} When one agent plans,
edits code, and declares completion in a single conversation, the human must reconstruct the goal, the changes,
and the supporting evidence from a long transcript or diff, and verification waits until both the
implementation and its review context have grown. \emph{Judgement engineering} instead places explicit,
mechanically enforced decisions at the boundaries between planning, implementation, review, and learning. Four
principles guide the design.

\begin{itemize}
\item \textbf{Judge the plan before the code.} A plan contract fixes the goal, acceptance criteria, scope, and
  tasks before execution, and gives the reviewer a stable basis for judgement.
\item \textbf{Separate building from judging.} The builder ends each round with a summary that links its claims
  to evidence and states what remains open; a different model decides whether the claims hold, and only it can
  declare completion (\S\ref{sec:alternation}).
\item \textbf{Decompose the completion decision.} Alignment review checks the claim against the plan and
  separates blockers from side issues that can be queued; code review checks the diff for defects. A change can
  follow the plan yet contain defects, or be locally correct yet miss the goal.
\item \textbf{Measure progress by evidence, not rounds.} A stalled loop should revisit the scope, escalate, or
  stop; a finished loop should leave lessons for later tasks.
\end{itemize}

These principles generalize the evaluator-optimizer pattern~\citep{anthropic2024buildingagents} from single
responses to whole working sessions, and \system{} keeps the orchestration itself in deterministic code rather
than in a model.

\subsection{Alternating Models as Joint Sampling}
\label{sec:alternation}

We view a coding session as a Markov chain over states $s_t$, the repository together with the plan and the
review record after round $t$. A single agent draws every step from one model, $s_{t+1} \sim P_M(\cdot \mid
s_t)$, and stops when the same model accepts its own state. Proposal and acceptance then share one set of blind
spots: whatever the model systematically gets wrong, it also tends to
accept~\citep{huang2024cannotselfcorrect,panickssery2024selfpreference}.

\begin{figure}[t]
  \centering
  \includegraphics[width=\linewidth]{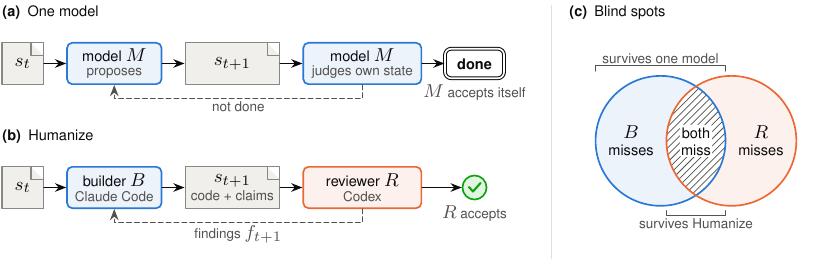}
  \caption{\textbf{Alternating models reduces accepted defects when their blind spots differ.}
    (a)~One model~$M$ proposes and judges each state, leaving its blind spots unchecked.
    (b)~\system{} forms a joint sampler: builder~$B$ proposes each state, and reviewer~$R$ from another vendor
    judges it and decides acceptance. (c)~A defect overlooked by $B$ survives the loop if $R$ also misses it.}
  \label{fig:alternation}
\end{figure}

\system{} alternates two models instead (Figure~\ref{fig:alternation}). The builder $B$ proposes the next state
given the reviewer's last findings, $s_{t+1} \sim P_B(\cdot \mid s_t, f_t)$, and a reviewer $R$ from another
vendor, in a fresh context, returns findings $f_{t+1} \sim P_R(\cdot \mid s_{t+1})$ that either accept the state
or condition the next proposal. Like a two-block Gibbs sampler, the chain alternates between two conditionals,
here supplied by two different models, so every step is shaped by both, and the chain can stop only in a state
that $R$ accepts. Let $b$ be the probability that the builder introduces a defect and overlooks it, and $m_R$
the probability that the reviewer misses it given that the builder did. The defect survives a single-model loop
with probability $b$ and the alternation with probability $b\,m_R$. Because $m_R$ approaches one as the
reviewer's blind spots approach the builder's, the joint chain accepts fewer defective states whenever the two
models' blind spots differ; this is why \system{} pairs Claude Code as builder with Codex as reviewer by
default. An external oracle, such as a proof checker or a test harness, adds a third factor, its own miss rate
$m_O$, and where it is complete for a claim it can stand in for the reviewer's independence
(\S\ref{sec:applications}). The price of alternation is that the reviewer's false positives also enter the
chain, as extra rounds (\S\ref{sec:findings}).

\section{System Design}
\label{sec:system}

\begin{figure}[t]
  \centering
  \includegraphics[width=\linewidth]{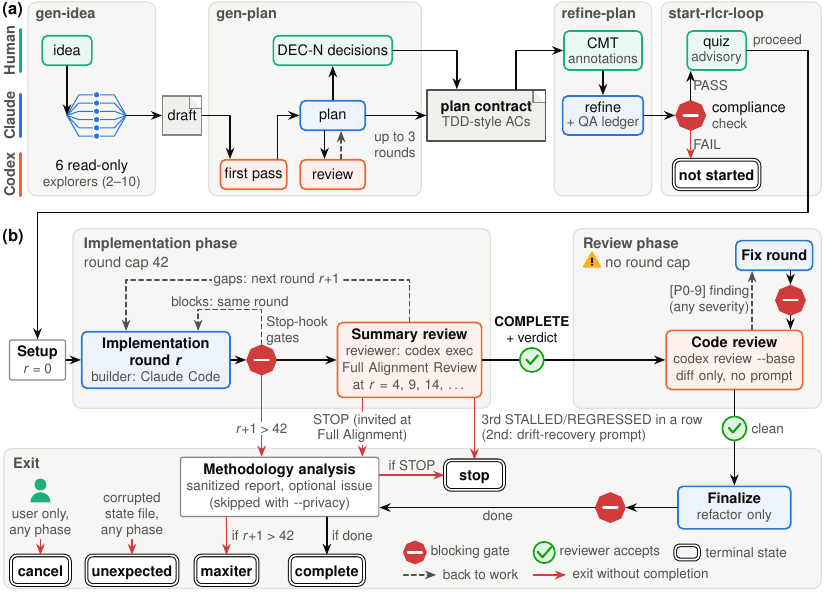}
  \caption{\textbf{\system{} fixes the plan before execution and makes completion a reviewer decision.}
    (a)~The planning chain separates human, Claude, and Codex roles. Lanes denote actors, regions denote
    commands, and the first three commands are optional. (b)~The RLCR state machine shows default transitions
    and explicit exits. Octagons mark blocking gates; dashed arrows return work; red arrows exit without
    completion.}
  \label{fig:rlcr-system}
\end{figure}

\system{} implements the workflow as a Claude Code plugin~\citep{anthropic2025plugins}. It provides five slash
commands, four subagents, six skills, and hooks for four Claude Code events~\citep{anthropic2025hooks}.
We describe the plugin's last release (Figure~\ref{fig:rlcr-system}).

\subsection{Planning as a Contract}
\label{sec:planning}

\system{} separates planning from implementation. \texttt{gen-idea} expands a rough idea into a draft
using parallel, read-only subagents that explore different directions. \texttt{gen-plan} turns this draft into
a structured plan, with coding forbidden during plan generation. Claude and Codex draft and critique the
plan for up to three rounds, then route unresolved points to the human. The plan is a contract, not a
disposable prompt: it defines the goal, acceptance criteria with positive and negative tests, path boundaries,
and tasks assigned to Claude for coding or Codex for analysis.

Before execution, the plugin runs two checks. A plan-compliance check rejects unrelated plans and
branch-switching instructions that would break loop accounting. An advisory quiz asks the human two
multiple-choice questions about the plan's mechanism and architecture. This added friction helps expose
plans accepted without being understood.

\subsection{The Round Protocol}
\label{sec:rlcr}

\begin{figure}[t]
  \centering
  \includegraphics[width=\linewidth]{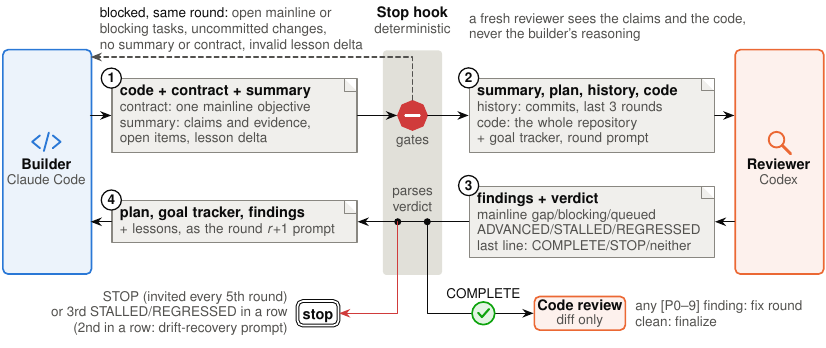}
  \caption{\textbf{\system{}'s Stop hook makes each RLCR round an evidence-based builder--reviewer exchange.}
    (1)~The builder submits code, a round contract, and a claim summary. (2)~After the hook's gates pass,
    a fresh reviewer checks the claims against the plan, history, and code. (3)~The reviewer returns findings
    and a verdict. (4)~The hook uses them to prompt the next round, start code review, or stop the loop.}
  \label{fig:round-protocol}
\end{figure}

The builder works in one Claude Code session, with rounds defined as ``One round = the agent believes the
entire plan is finished.'' The Stop hook mediates each exchange with the reviewer
(Figure~\ref{fig:round-protocol}). It supplies the plan, the goal tracker, and the previous review findings.
The goal tracker's immutable section holds the goal and acceptance criteria. The hook also points the
builder to BitLesson, whose selector chooses project lessons relevant to each task. The builder returns
commits, a round contract naming one mainline objective, and a summary of changes, supporting evidence,
and open work. The contract queues side issues so they do not take over the round.

When the builder tries to stop, the hook checks these artifacts before launching a fresh Codex process.
The reviewer reads the plan, recent rounds, code, and the builder's claims and evidence. It separates findings
into mainline gaps, blocking side issues, and queued side issues, and reports whether the mainline advanced,
stalled, or regressed. Returning \texttt{COMPLETE} requires a finished plan; deferred tasks count as incomplete.
The hook parses the verdict and turns the findings into the next round's prompt. Two consecutive stalled
rounds trigger a drift-recovery prompt. Every fifth round is a Full Alignment Review that audits all acceptance
criteria. Invitations to stop the loop are restricted to these reviews.

This exchange implements the alternation in \S\ref{sec:alternation}. The reviewer judges the repository state
and the builder's claims with evidence, not the builder's reasoning. The builder receives findings through a
prompt it cannot edit, and the hook parses the verdict.

Once the reviewer accepts the implementation, the loop enters code review. Codex checks the diff against the
base commit and emits severity-tagged findings. Any finding starts a fix round within this phase; a clean review
allows finalization through a behavior-preserving simplification pass. This phase has no round cap;
\S\ref{sec:findings} examines its cost.

The loop has five explicit terminal states: \texttt{complete}; \texttt{stop}, from the reviewer's verdict or
three consecutive stalled or regressed rounds; \texttt{maxiter}, at the implementation cap (42 rounds by
default); \texttt{cancel}, which requires a user request; and \texttt{unexpected}, for a corrupted state file.
After \texttt{complete}, \texttt{maxiter}, or a reviewer-requested \texttt{stop}, an analysis agent turns the
recorded rounds into a sanitized methodology report. The user may file it as an issue after two confirmations.
These reports form the corpus in \S\ref{sec:corpus}.

\subsection{Mechanical Gates}
\label{sec:gates}

\system{} enforces the workflow through code rather than relying on prompts. We catalogued 72 mechanical
gates (Appendix~\ref{app:gates}). Hook validators block builder writes to the state file, the plan copy, the
immutable goal, and other rounds' files. Before launching a review, the Stop hook checks that summaries,
lesson deltas, task lists, and commits are present. It also checks that review outputs are well formed.
These gates enforce artifact presence and form; the reviewer judges content. They keep the workflow mandatory
when an agent errs or over-claims. Enforcement grew with the system: the Stop hook expanded from 975 to 2,221
lines.

\paragraph{Runtimes.}
\label{sec:runtimes}
In the default runtime, the user's Claude Code session builds, and Codex reviews with \texttt{gpt-5.5}
at high effort. The plugin protects independence through a reviewer from another vendor, a fresh process per
review, hook-parsed verdicts, and prompts the builder cannot edit. Community-contributed Codex and Kimi
runtimes implement fewer of these protections. Host-tool dependencies also affect reliability: later Codex CLI
releases broke parts of the plugin after its last release.

\section{Deployment and Postmortem Study}
\label{sec:deployment}

\paragraph{Deployment.}
\system{} grew out of GAAC, an earlier Claude Code methodology plugin~\citep{liu2025gaac}. Its first release
contained the core loop. Over 108 days, 68 versions reached the main branch, adding code review, agent teams,
the Codex and Kimi runtimes, BitLesson, and joint plan generation by Claude and Codex. The last release added
round contracts, drift control, and the methodology analysis that generated most of the reports below.
Development then moved to a dev branch and a standalone successor~\citep{humanfia2026humanize2}.

Adoption continued after the plugin's last release. Its repository had 1,468 stars, with 61\% gained after
that release, 130 forks, and pull requests from 34 authors. Contributors outside the core team supplied
the Codex and Kimi runtimes, BitLesson, plan refinement, round contracts, and a monitoring dashboard.

\paragraph{Postmortem corpus.}
\label{sec:corpus}
We analyze \system{}'s public issue tracker as an observational corpus of methodology reports. It contained
150 issues from 52 authors, including 118 postmortems of real loops. Of these postmortems, 113 carry the
signature or template of the methodology analysis (\S\ref{sec:rlcr}).

LLM agents coded every issue using a written codebook (Appendix~\ref{app:corpus}). The codebook records issue
type, stated round count and exit, reported pressures, and proposed mechanisms. Agreement with a blind second
LLM coder reached Cohen's $\kappa = 0.93$ for round counts and $\kappa = 0.79$ for primary pressure.
Artifact state, review tails, scope, and convergence were the most frequent pressures, each present in
64--73\% of postmortems. Review tails and convergence were also the most common primary pressures.

The corpus is not a representative sample of all runs. Users chose what to file, four reporters wrote half
the reports, and the methodology analysis runs for selected exits. We use these reports to examine where
judgement boundaries held, where they were too weak, and when further review cost more than it helped.

\paragraph{Loop length.}
\label{sec:loop-lengths}
Of the 93 postmortems with a stated round count, the median loop ran 10 rounds and the longest ran 87.
In the 27 postmortems that separated rounds by phase, review dominated: 364 of 541 rounds (67.3\%) occurred
in the uncapped code-review phase, after implementation was accepted. Before acceptance, stagnation stops
followed the review schedule: every such stop with a stated round number fell on a Full Alignment Review,
where the reviewer is invited to stop the loop.

\section{Applications}
\label{sec:applications}

We next examine applications with checkable outcomes. They are application reports, not controlled
comparisons.

\paragraph{Loom: faster search with trace equivalence\protect\footnote{\url{https://github.com/PolyArch/loom}}.}
Loom's placement-and-routing mapper uses simulated annealing, so optimizing its search loop can change the
decisions it makes~\citep{liu2026loom}. The optimization contract required exact agreement with the reference
mapper's per-level decision trace. Under a 180-second budget, the campaign raised the explored proposal slots
from 97,138 to 227,777 (2.34$\times$), and four faster candidates that broke trace equivalence were reverted.
This is a throughput result, not evidence of better mappings.

\paragraph{Gem5: a build-system migration\protect\footnote{\url{https://github.com/gem5/gem5/pull/2969}}.}
Replacing gem5's SCons build must not change what the build generates. A loop on the \system{} plugin, with Claude
building and Codex reviewing, replaced it with CMake and Ninja and added a Bazel overlay across 567 files in 133
commits. The first author validated the result with the
full build, unit tests, and matching simulation objects across the two new builds; upstream, its CI has not
passed and the pull request awaits maintainer review. An agent loop can produce a change of this size; accepting
it remains a human decision.

\begin{figure}[t]
  \centering
  \includegraphics[width=\linewidth]{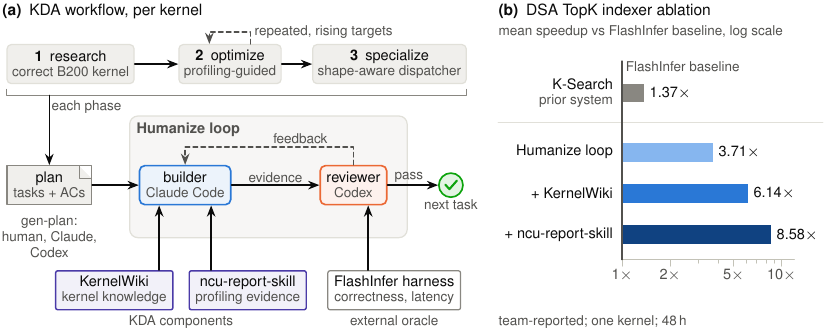}
  \caption{\textbf{Kernel knowledge and profiling feedback improve KDA's reported DSA indexer speedup.}
    (a)~The builder queries KernelWiki for production kernels, contest submissions, and documentation, and
    uses ncu-report-skill to interpret Nsight Compute profiles. The reviewer checks this evidence against each
    acceptance criterion, using the official FlashInfer harness as oracle. (b)~The team-reported 48-hour DSA
    TopK indexer ablation adds each component to the \system{} loop.
    K-Search is a separate prior system.}
  \label{fig:kda}
\end{figure}

\paragraph{Kernel Design Agents (KDA)\protect\footnote{\url{https://nvlabs.github.io/kda}}.}
KDA builds a domain-specific knowledge base and fast profiling feedback around the \system{}
loop~\citep{hanlab2026kda} (Figure~\ref{fig:kda}). KernelWiki is a retrieval store of production kernel pull
requests, contest submissions, and documentation that the builder queries while it designs and optimizes a
kernel; the ncu-report-skill turns Nsight Compute profiles into evidence that the builder acts on and the
reviewer checks against each acceptance criterion. Each kernel passes through three plans, for a correct kernel,
for profiling-guided optimization with rising speedup targets, and for shape-aware dispatch, and an RLCR loop
executes each plan with Claude as coding agent and Codex as acceptance agent. In the MLSys 2026 FlashInfer
contest~\citep{flashinfer2026contest}, KDA placed first in MoE, second in DSA, and third in GDN in the Full-Agent
division, with team-reported mean speedups over the FlashInfer baselines of 19.08$\times$ for the DSA indexer,
4.54$\times$ for DSA attention, and 1.92$\times$ for GDN prefill; two of its five kernels remained slower than the
baseline. A later team-reported ablation on the DSA indexer, under a 48-hour budget, credits each part: the loop
alone reached 3.71$\times$, KernelWiki raised it to 6.14$\times$, and the profiling skill to 8.58$\times$, against
1.37$\times$ for K-Search~\citep{cao2026ksearch}. In the same rerun, KDA exceeded K-Search on all three tracks:
11.9$\times$ against 3.76$\times$ for DSA, 1.16$\times$ against 1.04$\times$ for GDN, and 0.72$\times$ against
0.27$\times$ for MoE, in mean speedup over the FlashInfer baselines.

\paragraph{Humanize Olympiad Agents (HOA)\protect\footnote{\url{https://github.com/humanfia/hoa}}.}
Humanfia's HOA applies the builder-reviewer workflow to olympiad and benchmark
problems~\citep{humanfia2026hoa}, with a completion check for each discipline.
Two runs proved all six IMO 2026
problems in Lean from AxiomProver's formal statements~\citep{axiommath2026imo}, one with GPT-5.6
Sol in both roles and one with Kimi K3 building and Codex reviewing. All six IOI 2026 tasks received full scores
from the Codeforces judge. Both IPhO 2026 theory runs scored 30/30 against the official marking scheme, IChO 2026
reached 418.5/437 theory points with all 68 formalization targets covered. The IBO 2024 run matched all
100 official answers. We use 2024 because the committee keeps problems confidential for two years.
These scores are self-graded; IBO agents could see the key. The releases share specialized agent flows with a
judgement boundary: preserve questions, check artifacts, and identify who accepted them.

\paragraph{PutnamBench\protect\footnote{\url{https://trishullab.github.io/PutnamBench/leaderboard}} and Lean-Eval\protect\footnote{\url{https://lean-lang.org/eval/legacy}}.}
\label{sec:app-math}
For formal proofs, the Lean kernel and a statement-preservation check run before the reviewer, which must also
obtain an independent proof verifier's approval. \system{} achieves 672/672 on
PutnamBench~\citep{humanfia2026putnambench}. Lean-Eval requires submitted formal proofs to pass its Comparator
check, and \system{} ranks first on its legacy leaderboard, ahead of professional teams including AxiomProver (Axiom Math),
Aleph Prover (Logical Intelligence), Aristotle (Harmonic), and Seed Prover (ByteDance)~\citep{leanfro2026leaneval}.

\begin{figure}[t]
  \centering
  \includegraphics[width=\linewidth]{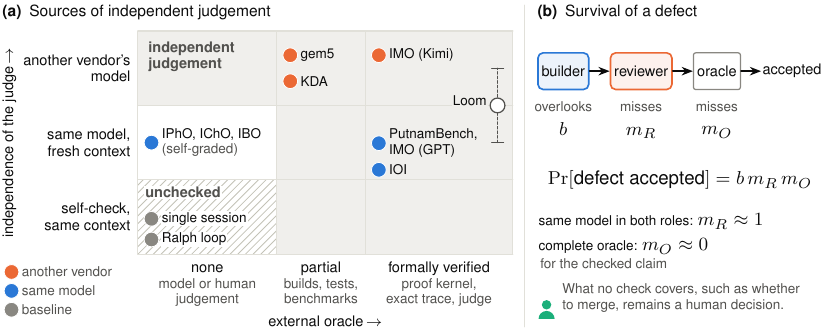}
  \caption{\textbf{Independent judgement can come from a second model or an external oracle.}
    (a)~Applications and baselines are positioned by reviewer independence and oracle strength; acceptance
    is unchecked when both are absent. (b)~A defect is accepted with probability $b\,m_R\,m_O$.
    A reviewer from another vendor can lower $m_R$; a complete oracle drives $m_O$ toward zero for the claim
    it checks.}
  \label{fig:independence}
\end{figure}

\paragraph{Two sources of independent judgement.}
The applications differ in where their independent judgement comes from (Figure~\ref{fig:independence}). Where
external checks are partial, as the builds and tests of the gem5 migration or the contest harness of KDA, a
reviewer from another vendor carries the judgement that the checks leave open, and the gem5 change still awaits
a human decision. Where an oracle is complete for the claim, as the Lean kernel for PutnamBench and IMO or the
judge for IOI, one model in both roles sufficed, because the oracle drives $m_O$ (\S\ref{sec:alternation})
toward zero for the property it checks. Results with neither source are the weakest: the natural-language olympiad scores come from HOA's own
grading and are estimates, and the single-session agent and the Ralph loop sit in this unchecked corner by
construction. The applications show what such loops can reach, not how much the reviewer adds: no public
comparison runs the same builder with and without the reviewer under the same budget, and within a run the loop
often confirms a first attempt, as in the PutnamBench rerun where 86 of 98 problems passed at the first turn.

\section{Findings}
\label{sec:findings}

Together, the postmortems and the applications yield five lessons for orchestrating coding agents.

\paragraph{Independent review is trusted but expensive.}
The most consistent lesson is that the reviewer, not the builder, should decide completion. Forty-five of
the 118 postmortems describe builders claiming more than they had done, and KDA's team reports the same failure
in kernel work: its builder once used its own first kernel as the baseline and declared the task done. Few
reports fault the reviewer's verdicts; the complaints concern how review is paced. A one-finding-per-round
cadence turns valid review into a long tail, and the code reviewer, given only the diff and no history,
rediscovers related defects one path at a time. Review should therefore be batched, not weakened:
group related findings and give the reviewer memory of earlier rounds, without importing the builder's
reasoning.

\paragraph{Iteration does not guarantee convergence.}
The review phase was exempt from the round cap; in the postmortems that split phases, two thirds of
rounds came after acceptance (\S\ref{sec:loop-lengths}); one ran 87 rounds against a cap of 42. A mainline
verdict such as ``advanced'' is useful only if it is tied to measurable movement toward the acceptance criteria,
and one report recorded ``advanced'' in every round of a loop that ended with none of its eight criteria fully
met. Termination must be designed: every phase needs a bound, a stop rule should follow the state of the work
rather than a fixed schedule, and the loop needs an exit for work that is blocked rather than failing.

\paragraph{Plans and scope are judgement boundaries.}
Large plans create too many simultaneous obligations, vague acceptance criteria produce ambiguous stopping
conditions, and reviewers discover real issues whose relation to the accepted plan is unclear. Builder and reviewer must also read the contract the same way: the plugin's rule that
deferred work counts as incomplete clashed with plans that deliberately left work for later. Discoveries should
be routed rather than ignored: a blocking issue should explain why it prevents the accepted goal, a queued issue
should be recorded without extending the loop, and a plan amendment should be explicit. Enforcement presupposes a
plan that the human can judge.

\paragraph{Evidence and environment decide termination.}
Several sessions stalled not because the implementation was wrong but because evidence could not be collected
consistently: acceptance criteria depended on unavailable infrastructure, or the builder's and reviewer's
environments disagreed. Verification is therefore part of the judgement design, not a test-running detail. The
strongest application results rest on oracles outside both agents, which can replace a second model as the
independent judge (\S\ref{sec:applications}). A robust loop records its environment assumptions and asks the
human for a decision instead of spending rounds on evidence it cannot collect.

\paragraph{The loop needs judgement around it.}
The methodology analysis made reports cheap: the 118 postmortems contain 771 numbered suggestions. No
postmortem proposal is strongly linked to a change on the plugin's main branch, whereas concrete bug reports were fixed within
days; 86 of the 150 issues arrived after the plugin's last release, when development had moved to the dev branch and
a standalone successor. Self-generated feedback scales reporting more easily than learning: a loop that judges code also
needs a process that judges its own reports.

\section{Related Work}
\label{sec:related}

\paragraph{Coding agents and loops.}
Tool-using agents interleave reasoning and action~\citep{yao2023react}, operate on whole
repositories~\citep{yang2024sweagent,wang2025openhands}, and refine their output with their own critique or with
execution feedback~\citep{shinn2023reflexion,madaan2023selfrefine,chen2024selfdebug}. The Ralph loop reruns such
an agent on the same prompt until it emits a completion promise~\citep{huntley2025ralph,anthropic2025ralphwiggum};
\system{} keeps the outer loop but gives the stop decision to an independent reviewer, and it adds an enforced
review to the explicit plans of spec-driven tools~\citep{delimarsky2025speckit,swaminathan2025kiro}. Vendors
have since released native loops, such as Claude Code's \texttt{/goal} with a separate
evaluator~\citep{anthropic2026goal}, and loop design has become a research
topic~\citep{li2026loopsbench,lulla2026loopengineering}; none of these studies evaluates \system{}.

\paragraph{Multi-agent orchestration.}
Multi-agent frameworks organize models into software roles through standard operating procedures, chat chains,
or programmable conversations~\citep{hong2024metagpt,qian2024chatdev,wu2024autogen}. A taxonomy of their failures
groups them into system design, inter-agent misalignment, and task verification~\citep{cemri2025mast}. \system{}
is deliberately small: two model roles and a human, with routing held by deterministic hooks rather than by a
model, and with verification assigned to a dedicated role.

\paragraph{Models as judges.}
Separate verifiers and critics catch errors that generators miss~\citep{cobbe2021verifiers,mcaleese2024criticgpt},
and models are widely used to judge model output~\citep{zheng2023judging,zhuge2025agentasajudge}. Critics also
report bugs that do not exist~\citep{mcaleese2024criticgpt}, and cross-model code review is asymmetric: in a
controlled study, Codex reviewing Claude's drafts, \system{}'s default pairing, was counterproductive when the reviewer could not
run tests~\citep{xiang2026crossmodelreview}. \system{}'s reviewer is an agent that can execute code and judges
whole sessions; the review tails of \S\ref{sec:loop-lengths} are one cost of that design.

\section{Limitations}
\label{sec:limitations}

\system{} depends on other vendors' command-line tools. Its orchestration runs on Claude Code's hook
interface and its reviewer on Codex CLI, whose releases after the plugin's last release broke parts of it, and only the
Claude Code plugin installs the full set of independence protections (\S\ref{sec:runtimes}). Its
termination rules are coarse: the round cap binds only the implementation phase, the reviewer is invited to stop
the loop only every fifth round, and no exit separates work blocked by its environment from work that fails
(\S\ref{sec:findings}). Its gates check the presence and form of artifacts rather than their content, so a
well-formed but false summary must be caught by the reviewer, and the tool-call validators fail open on a
malformed state file (Appendix~\ref{app:gates}). The code reviewer sees only the diff, without memory of earlier
rounds, and \system{} keeps no telemetry on gate firings or round costs. Finally, \system{} has not been evaluated in a
controlled comparison: the evidence in this paper is observational, and no public study runs the same builder
with and without its reviewer.

\section{Conclusion}
\label{sec:conclusion}

\system{} treats judgement as the primary design problem of agentic coding. Its contribution is not a larger
agent organization or a stronger claim about autonomous coding, but a compact multi-agent workflow that
alternates two models as builder and reviewer, makes goals, claims, reviews, evidence, and stopping decisions
explicit, and enforces them mechanically. In deployment,
independent review caught over-claiming builders, and the builder-reviewer pattern supported applications from a
build-system migration to a complete PutnamBench proof set, but iteration did not guarantee convergence, and
cheap reports still needed judgement to yield changes. Reliable agentic coding requires not just generation but
engineered judgement around it. Development continues in a standalone successor, which moves the loop out of host-CLI hooks
into its own runtime~\citep{humanfia2026humanize2}; we will describe it separately.

\subsubsection*{Acknowledgments}

We thank the contributors to the first major version of \system{} for the following:
\begin{itemize}[itemsep=1pt,topsep=3pt]
\item {Codex and Kimi runtimes:} \mbox{\href{https://github.com/shinezyy}{Yaoyang Zhou}}.
\item {Plan convergence, bitLesson, and plan refinement:} \mbox{\href{https://github.com/ZenusZhang}{Dongyu Zhang}}.
\item {Round contracts and drift control:} \mbox{\href{https://github.com/tastynoob}{Yan Xu}}.
\item {Integral review context and dashboard:} \mbox{\href{https://github.com/zevorn}{Chao Liu}}.
\item {Idea generation:} \mbox{\href{https://github.com/shinan6}{Shinan Liu}}.
\item {Roadmap discussions and infrastructure support:} \mbox{\href{https://entropy-xcy.github.io/}{Ceyu Xu}}.
\item {SGLang integration and infrastructure discussion:} \mbox{\href{https://github.com/bbuf}{BBuf Xiaoyu Zhang}}.
\item {HAN Lab Kernel Mafia on the KDA contest:} \href{https://dongyunzou.github.io/}{Dongyun Zou},
\href{https://scholar.google.com/citations?user=aYh01A4AAAAJ}{Zhekai Zhang},
\href{https://ys-2020.github.io/}{Shang Yang}, \href{https://haokang.me/}{Hao Kang},
\href{https://muyangli.com/}{Muyang Li}, \href{https://han-cai.github.io/}{Han Cai},
and \href{https://lzhu.me/}{Ligeng Zhu}.
\item {Olympiad Competition, PutnamBench and Lean-Eval Leaderboard:} \href{https://zhengyangzhang06.github.io/}{Zhengyang Zhang}, \href{https://menik1126.github.io/}{Jing Xiong},
\href{https://github.com/hongzhoulin89}{Hongzhou Lin},
\href{https://web.math.princeton.edu/~jc1220/}{Jui-Hui Chung}, and \href{https://lzhu.me/}{Ligeng Zhu}.
\item {FlashInfer team: }
\href{https://scholar.google.com/citations?user=GcJAM44AAAAJ}{Yixin Dong},
\href{https://scholar.google.com/citations?user=AI8jzA8AAAAJ}{Hongyi Jin},
\href{https://expye.com/about}{Zihao Ye},
\href{https://scholar.google.com/citations?user=9xQ47MUAAAAJ}{Avery Huang},
and \href{https://scholar.google.com/citations?user=Nq0CpwYAAAAJ}{Junru Shao},
\item The MLSys kernel contest organizers, the PutnamBench and Lean-Eval maintainers.
\end{itemize}

also including everyone who opened a pull request or filed an issue or shared an idea during discussion. The main development is led by \href{https://www.nvidia.com/en-us/research}{NVIDIA Research}, while we appreciate the development and compute support from the \href{https://polyarch.cs.ucla.edu/}{UCLA PolyArch},
\href{https://hanlab.mit.edu/}{MIT HAN Lab}, and \href{https://catalyst.cs.cmu.edu/}{CMU Catalyst}.

\bibliographystyle{iclr2027_conference}
\bibliography{references}

\newpage
\appendix
\section{Mechanical Gates}
\label{app:gates}

We count as a gate every point where the code of \system{} checks the loop and
can block, end, or redirect it; Table~\ref{tab:gates} groups the 72 gates by category. Twenty-nine live in the
tool-call and prompt validators, which only the Claude Code plugin installs, and 32 run in the Stop hook. One
gate fails open: on a malformed state file, the tool-call validators exit with code~1, which Claude Code treats
as a non-blocking error~\citep{anthropic2025hooks}, so the call proceeds.

\begin{table}[h]
  \caption{\textbf{The largest gate categories protect state, plans, and round isolation.} All 72 gates of
    \system{} by category.}
  \label{tab:gates}
  \centering
  \footnotesize
  \begin{tabularx}{\linewidth}{@{}>{\raggedright\arraybackslash}p{3.4cm}r>{\raggedright\arraybackslash}X@{}}
    \toprule
    Category & Gates & What the gates enforce \\
    \midrule
    State integrity & 12 & read-only loop state; schema, branch, and corruption checks \\
    Plan and scope protection & 11 & plan and compliance checks at setup; frozen plan copy and goal \\
    Round isolation & 10 & each round writes only its own files; other rounds are unreadable \\
    Git hygiene & 9 & a clean tree at setup and at every stop; loop state is never committed \\
    Review protocol & 7 & valid review and verdict each round; STOP and drift breakers; round cap \\
    Session hygiene & 6 & a locked-down analysis phase; no direct hook runs; only the user cancels \\
    Artifact presence & 5 & an initialized goal tracker; a contract and a summary before each review \\
    Lesson memory & 5 & every summary records a valid BitLesson delta \\
    Environment precondition & 4 & Codex, git, and a base branch are available \\
    Completion discipline & 3 & no open mainline or blocking tasks; no changed file over 2,000 lines \\
    \bottomrule
  \end{tabularx}
\end{table}

\section{Corpus Coding}
\label{app:corpus}

The corpus holds the 150 issues of PolyArch/humanize: 118 postmortems and 32 bug reports, feature requests,
questions, and other issues. For each postmortem, LLM agents coded the stated round count (93 postmortems), the
exit (complete 37, stagnation stop 21, maxiter 10, cancel 7, drift stop 1, not stated 42), the split of rounds by
phase (27), the pressures (Table~\ref{tab:corpus}), and the proposed mechanisms (771 numbered suggestions);
counts inside an issue are self-reports. A blind second LLM coder recoded a stratified sample of 30 postmortems
mixed with 6 decoys and agreed with Cohen's $\kappa = 0.93$ on round counts, $0.82$ on exits, and $0.79$ on the
primary pressure.

\begin{table}[h]
  \caption{\textbf{Review tails and convergence are the most common primary pressures.} Counts over the 118
    postmortems; a report can show several pressures but has one primary pressure. $\kappa$: agreement with the
    second coder.}
  \label{tab:corpus}
  \centering
  \footnotesize
  \begin{tabularx}{\linewidth}{@{}>{\raggedright\arraybackslash}p{2.5cm}>{\raggedright\arraybackslash}X rrr@{}}
    \toprule
    Pressure & Observed in the run & Present & Primary & $\kappa$ \\
    \midrule
    artifact state & tracker, summaries, contracts, lessons, review files & 86 & 3 & 0.71 \\
    review tail & findings one at a time; piecemeal fixes; oscillation & 80 & 36 & 0.77 \\
    scope and plan & size or ambiguity of the plan; scope creep; deferrals & 79 & 16 & 0.82 \\
    convergence & no stop when due, or a stop without converging & 76 & 27 & 0.70 \\
    evidence & environments, external jobs, test adequacy, sandbox & 58 & 19 & 0.93 \\
    over-claiming & claims the evidence does not support & 45 & 8 & 0.93 \\
    cost & tokens, context, compute, or time beyond rounds & 25 & 1 & 0.71 \\
    human escalation & decisions that needed a human but were not routed & 24 & 3 & 1.00 \\
    runtime and tooling & CLIs, hooks, operating systems, setup & 13 & 4 & 0.63 \\
    other & & 4 & 1 & -- \\
    \bottomrule
  \end{tabularx}
\end{table}

\end{document}